\documentclass{ecai} 
\usepackage{latexsym}
\usepackage{amssymb}
\usepackage{amsmath}
\usepackage{amsthm}
\usepackage{booktabs}
\usepackage{enumitem}
\usepackage{graphicx}
\usepackage{color}
\usepackage{subcaption}
\usepackage[normalem]{ulem}

\newcommand{\BibTeX}{B\kern-.05em{\sc i\kern-.025em b}\kern-.08em\TeX}

\begin{document}

%%%%%%%%%%%%%%%%%%%%%%%%%%%%%%%%%%%%%%%%%%%%%%%%%%%%%%%%%%%%%%%%%%%%%%%%

\begin{frontmatter}

%%% Use this command to specify your submission number.
%%% In doubleblind mode, it will be printed on the first page.

\paperid{123} 

%%% Use this command to specify the title of your paper.

\title{Performance-Efficiency Trade-off for Fashion Image Retrieval}

%%% Use this combinations of commands to specify all authors of your 
%%% paper. Use \fnms{} and \snm{} to indicate everyone's first names 
%%% and surname. This will help the publisher with indexing the 
%%% proceedings. Please use a reasonable approximation in case your 
%%% name does not neatly split into "first names" and "surname".
%%% Specifying your ORCID digital identifier is optional. 
%%% Use the \thanks{} command to indicate one or more corresponding 
%%% authors and their email address(es). If so desired, you can specify
%%% author contributions using the \footnote{} command.

%\author[A]{\fnms{First}~\snm{Author}\orcid{....-....-....-....}\thanks{Corresponding Author. Email: %somename@university.edu.}\footnote{Equal contribution.}}
%\author[B]{\fnms{Second}~\snm{Author}\orcid{....-....-....-....}\footnotemark}
%\author[B,C]{\fnms{Third}~\snm{Author}\orcid{....-....-....-....}} 

%\address[A]{Short Affiliation of First Author}
%\address[B]{Short Affiliation of Second Author and Third Author}
%\address[C]{Short Alternate Affiliation of Third Author}

\author[A]{\fnms{Julio}~\snm{Hurtado}\thanks{Corresponding Author. Email: julio.hurtado@warwick.ac.uk}\footnote{Equal contribution.}}
\author[A]{\fnms{Haoran}~\snm{Ni}\footnotemark}
\author[A]{\fnms{Duygu}~\snm{Sap}}
\author[B]{\fnms{Connor}~\snm{Mattinson}} 
\author[C]{\fnms{Martin}~\snm{Lotz}} 

\address[A]{CAMaCS, University of Warwick}
\address[B]{TRUSS}
\address[C]{Mathematics Institute, University of Warwick}

%%% Use this environment to include an abstract of your paper.

%\author{%
%Julio Hurtado\\
%University of Warwick\\
%\email{julio.hurtado@warwick.ac.uk}\And
%Haoran Ni\\
%University of Warwick\\
%\email{haoran.ni@warwick.ac.uk}\And
%Duygu Sap\\
%University of Warwick\\
%\email{duygu.sap@warwick.ac.uk}\And
%Connor Mattison\\
%TRUSS\\
%\email{connor@trussarchive.com}\And
%Martin Lotz\\
%University of Warwick\\
%\email{martin.lotz@warwick.ac.uk}
%}

\begin{abstract}
% haoran version:
The fashion industry has been identified as a major contributor to waste and emissions, leading to an increased interest in promoting the second-hand market. Machine learning methods play an important role in facilitating the creation and expansion of second-hand marketplaces by enabling the large-scale valuation of used garments. We contribute to this line of work by addressing the scalability of second-hand image retrieval from databases. By introducing a selective representation framework, we can shrink databases to 10\% of their original size without sacrificing retrieval accuracy. We first explore clustering and coreset selection methods to identify representative samples that capture the key features of each garment and its internal variability. Then, we introduce an efficient outlier removal method, based on a neighbour-homogeneity consistency score measure, that filters out uncharacteristic samples prior to selection. We evaluate our approach on three public datasets: DeepFashion Attribute, DeepFashion Con2Shop, and DeepFashion2. The results demonstrate a clear performance-efficiency trade-off by strategically pruning and selecting representative vectors of images. The retrieval system maintains near-optimal accuracy, while greatly reducing computational costs by reducing the images added to the vector database. Furthermore, applying our outlier removal method to clustering techniques yields even higher retrieval performance by removing non-discriminative samples before the selection.
\end{abstract}

\end{frontmatter}

%%%%%%%%%%%%%%%%%%%%%%%%%%%%%%%%%%%%%%%%%%%%%%%%%%%%%%%%%%%%%%%%%%%%%%%%

\section{Introduction}

The fashion industry is one of the largest producers of $CO_2$ due to the constant manufacturing of clothing and use of transportation. This issue has worsened with the rise of fast fashion \cite{li2024carbon}, emphasising the purchase of typically low-quality garments worn only a few times before being discarded. One alternative to this trend is to promote the reuse of clothing by encouraging people to wear items multiple times or by facilitating the buying, selling, or exchanging of used garments \cite{FastFashionWeb}.

To encourage the reuse of garments, some companies have focused on creating platforms for the second-hand market, which has sprung up the buying and selling of used goods. Their main goal is to minimise the use of resources in making new garments and minimise the transport costs, as these companies often focus on local markets.

\begin{figure}[t]
  \centering
   \includegraphics[width=0.85\linewidth]{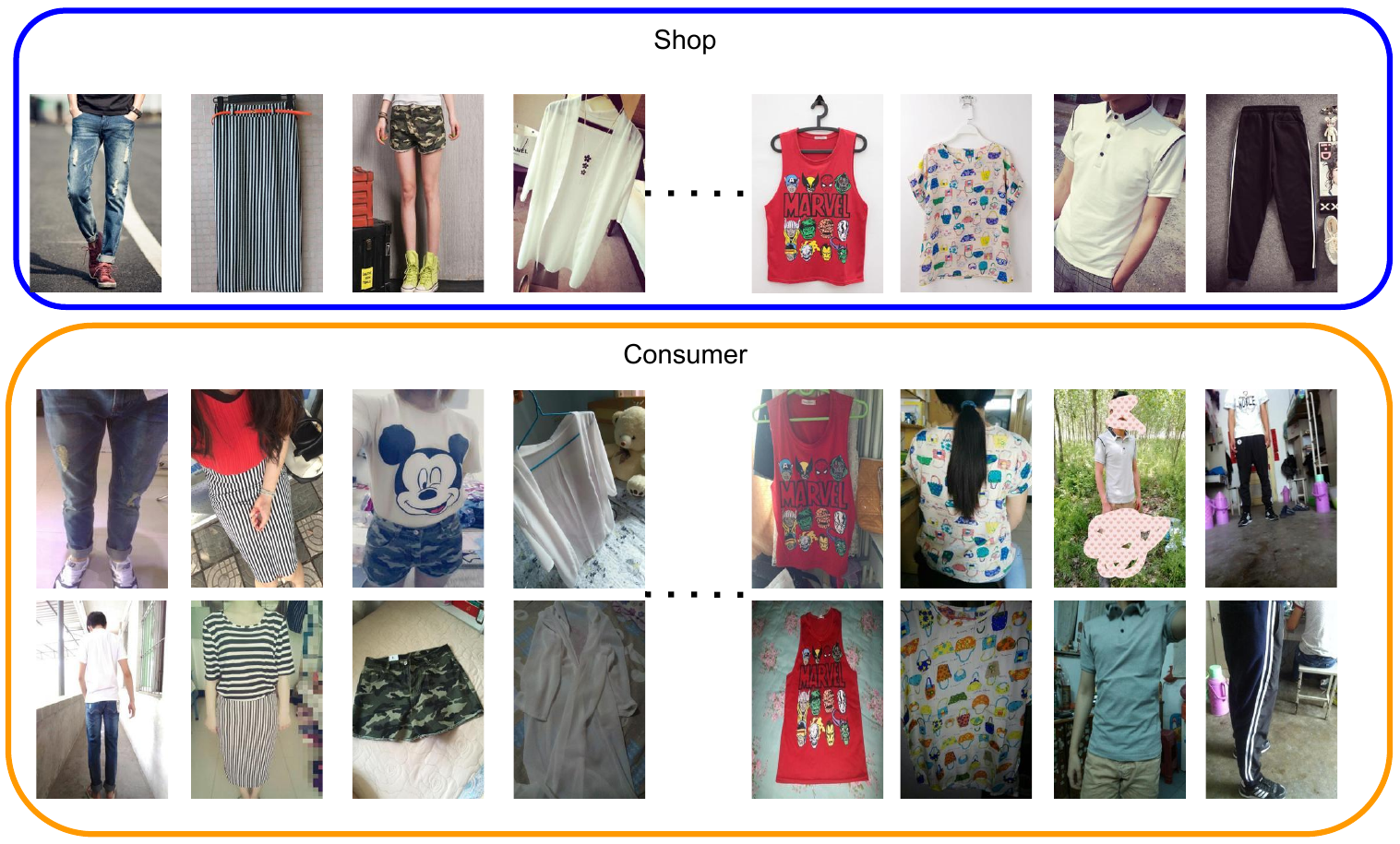}
   \hspace{5mm}
   \caption{This is an example of intra-class diversity in the Deep Fashion Dataset \cite{liuLQWTcvpr16DeepFashion}. The top line shows an example of a unique item uploaded by a shop. The rows below show images of the same item uploaded by consumers. These images exemplify the diversity among items of the same garment, where an object may have more visual similarities with objects from another garment than of the same category.}
   \hspace{150mm}
   \label{fig:diversity_garms}
\end{figure}

As the market increases, sellers face one significant challenge: determining the correct price for the products they wish to sell. The large variety of garments with similar characteristics makes it challenging to find visually comparable items, forcing them to conduct lengthy manual data analysis. This issue is further complicated by the significant price variations within the same category of garments, primarily due to the freedom sellers have when listing their products. Consequently, while there are many similarities among different products, there are also considerable differences within the same product type, as shown in Figure \ref{fig:diversity_garms}. The significant internal variations within a fashion category and the notable similarities across the categories complicate the retrieval of similar products.

Currently, the best-performing methods to retrieve similar items consist of using pre-trained Deep Learning models to generate visual feature vectors and create a vector database \cite{Chia2022, Marqo_fashionclip}. Then, when a new image arrives, using the same model, a vector is created and compared against the whole dataset. Some previous works focused on fine-tuning models for the fashion domain, with publicly available datasets \cite{liu2016mvc, liuLQWTcvpr16DeepFashion, DeepFashion2, han2017automatic}.

The dynamism of the working environment makes it difficult to rely on straightforward classification models, as the cost of training the model with new and old images is extremely high, both computationally and environmentally. Thus, large vector databases remain the most reliable option, since when new products arrive, new vectors are simply added to the database, instead of constantly retraining a classification model. However, vector databases cause another problem: the incremental cost of comparing a new image against the growing database.

Including all available images in the database enhances retrieval performance. However, this also increases the computational cost proportionally to the number of images added.
%However, the high cost makes it counterproductive to focus on reducing $CO_2$ emissions in the fashion market to generate more with the solution. 
Therefore, in this paper, we focus on proposing alternatives to adding all images to the database and propose adding a subset of representatives of each garment. Our proposal helps in two directions: (1) identify key concepts of a garment to differentiate it from other similar garments and (2) identify sub-distributions within a garment to represent the internal variability of a garment.

We study methods based on clustering \cite{sener2018active} and coresets~\cite{guo2022deepcore} to find samples that are representative of the garment. These methods focus on finding sub-distributions within the whole garment distribution, which are then used to select relevant samples to populate the database. These methods allow us to achieve similar performance when selecting only 10\% of the whole dataset. However, our study shows that selection does not improve retrieval performance. A closer look reveals that the large variability within each category is detrimental to the performance of clustering-based methods. To address this, we propose a method to remove non-discriminative samples based on the consistency score (C-Score) \cite{jiang2021characterizing}, which allows us to remove noisy elements and find better representatives.

The rest of the paper is divided as follows: first, we explore previous work in image retrieval in the fashion domain and different selection alternatives. We then explain how the selection of representations can help reduce the retrieval cost. In a next step, we show how to remove these noisy elements from the database. Finally, we show the results of the proposal on different public datasets focused on the fashion second-hand market.

%%%%%%%%%%%%%%%%%%%%%%%%%%%%%%%%%%%%%%%%%%%%%%%%%%%%%%%%%%%%%%%%%%%%%%%%

\section{Related Work}

The evolution of image retrieval methods within the fashion domain reflects advances in dataset and annotation quality. Traditional systems predominantly used content-based or attribute-based retrieval \cite{dubey2021decade}, where user queries were limited to selecting predefined filters or keywords. Such approaches, while effective to an extent, are often too rigid to capture the subtle differences that define fashion styles and user preferences. 

With the emergence of deep convolutional neural networks \cite{he2016deep}, retrieval systems have evolved to learn robust feature representations that capture the underlying visual semantics of fashion items. More recently, transformer-based architectures have been employed to fuse multi-modal information, combining image features with natural language feedback and attribute cues \cite{radford2021learning, liu2023visual}. This multimodal approach enables systems to handle complex queries beyond simple attribute matching; for instance, a user might request “a red, sleeveless dress with a lower neckline” and expect the system to retrieve images that accurately match this composite description. Pre-trained models such as FashionCLIP \cite{Chia2022} and vision transformers (ViT) \cite{dosovitskiy2020image} have been instrumental in this evolution, with Marqo’s fine-tuning of FashionCLIP further enhancing retrieval performance by leveraging large-scale, fashion-specific pre-training \cite{Marqo_fashionclip}. These models serve as robust backbones, providing rich, joint embeddings of visual and textual data, and are key to building interactive, high-performance fashion image retrieval systems that meet the nuanced needs of modern applications.

While the progress in fashion image retrieval has been impressive, the efficiency of training and deploying these complex models remains a critical challenge. As datasets continue to grow in size, training deep networks becomes computationally expensive and time-consuming. Coreset selection \cite{guo2022deepcore}, as a technique that seeks to identify and retain a small, representative subset of the training data, has emerged as a promising solution to this problem. By selecting the most informative samples from a large dataset, coreset methods enable efficient model training without a significant loss in performance. Traditional coreset methods were based on geometric \cite{chen2010super} or clustering techniques \cite{sener2018active}, but recent research has shifted towards more sophisticated approaches, including uncertainty-based \cite{coleman2020selection} and gradient matching methods \cite{killamsetty2021grad}. These methods help reduce training times and computational costs, which is particularly valuable for tasks such as neural architecture search and continual learning. 

In summary, the synergy between advanced fashion datasets, state-of-the-art image retrieval techniques, and coreset selection methods is an unexplored path that can represent a robust framework for building highly efficient and effective fashion image retrieval systems.

\section{Methodology}
\label{sec:metho}

\subsection{Problem formulation}

The fashion industry is constantly evolving, with new garments, colours, and styles emerging regularly. The second-hand market reflects this dynamism and is even more diverse due to the availability of old, new, and modified clothing options. This ever-changing landscape makes it difficult to categorise images because of the complexity involved in searching for specific items.

Many machine learning methods operate on the assumption of an independent and identically distributed (i.i.d.) dataset for model training. However, this assumption is violated when data from new groups emerge during inference. One solution is continuously retraining the model whenever new garment classes are introduced. Nevertheless, the high dynamism combined with the substantial costs associated with ongoing retraining encourages us to explore alternative approaches. Additionally, despite significant advances in the field of Continual Learning \cite{hurtado2023continual, wang2024comprehensive}, there remain challenges in training models continuously without experiencing forgetting.

For this reason, and similar to previous studies \cite{zhu2024generalized}, we propose to address this problem as an image retrieval task. Informally, we can define the setup as follows: starting with an initial set of training images, we create a database of feature vectors generated from a pre-trained model. When a new image arrives, we can select the most similar vectors using some similarity metric (e.g. cosine similarity) and label that image based on the top-k most similar samples \cite{waqas2024exploring}.

Although applying image retrieval is a feasible solution, retrieving costs increase linearly as we increase the number of samples in the training set, as shown in Figure \ref{fig:elapsed_time}. Solutions based on Hashing \cite{luo2023survey} have been proposed to decrease the retrieval and storage cost; however, this paper will present an alternative that complements previous works. We analyse and propose the creation of a subset of examples that best represent the complete dataset. As detailed below, this proved challenging due to the significant percentage of noisy elements.

\subsection{Garment representatives}
\begin{figure*}
  \centering
  \begin{subfigure}{0.32\linewidth}
     \includegraphics[width=0.9\linewidth]{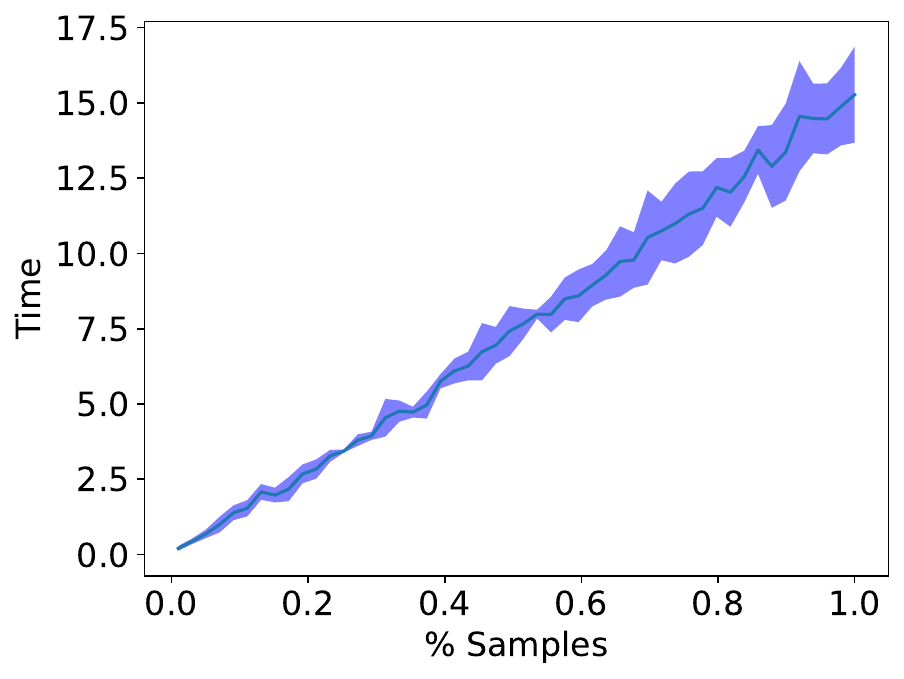}
    \caption{Time elapsed}
    \label{fig:elapsed_time}
  \end{subfigure}
  \hfill
  \begin{subfigure}{0.64\linewidth}
    \includegraphics[width=0.9\linewidth]{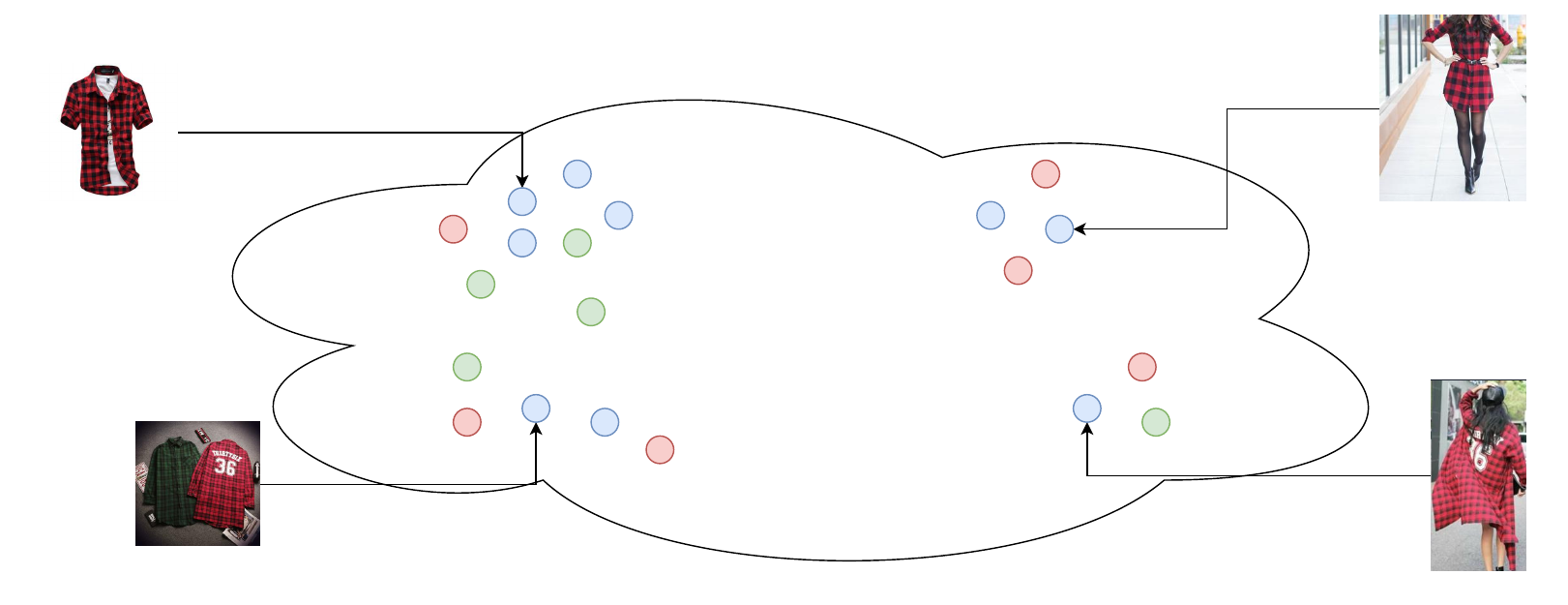}
    \caption{Vector Space with different views}
    \label{fig:vector_space}
  \end{subfigure}
  \hspace{15mm}
  \caption{In most cases, accuracy and computational efficiency are trade-offs during retrieval. A few samples representing the garment can decrease the computational resources needed to retrieve the most similar items, as shown in (a). However, as shown in (b), it is challenging to represent a garment with few samples fully.}
  %\label{fig:representation_space}
\end{figure*}

Typically, more images make representing the vast distribution space easier. However, ideally, one image per garment should be enough to describe the garment fully in the vector dataset. In scenarios with high input space diversity, like in the second-hand market, selecting only one representative per class becomes challenging. As shown in Figure \ref{fig:diversity_garms}, consumers and shops—whether companies or individuals—upload a wide variety of images of the same garment using different background colours, lighting, and quality. 

The high diversity of images poses a challenge in representing the entire set with a single garment, as no single image can capture all key features, especially when there is no training or fine-tuning. The occlusion of important details further complicates this issue; for instance, both the front and back of a t-shirt are essential for identifying its unique characteristics when unique patterns are involved. This leads to the question: How many images should we select to minimise costs while maximising retrieval accuracy?

One way to approach the problem is by considering the representation space where retrieval occurs. Assuming we have a pre-trained model $F_{\theta}$, we can generate a feature vector $v_i$ of each image $x_i$ using $v_i = F_{\theta}(x_i)$, which is usually generated by extracting the representation just before the linear classifier. Depending on the training objectives of the model, these vectors may emphasise various characteristics, resulting in a diverse set of sub-distributions, as illustrated in Figure \ref{fig:vector_space}. Here, the right and left sides are different because of the inclusion (or exclusion) of a person in the image. 

The previous observation suggests that different internal groups for each garment can be identified, allowing us to select a representative from each group to create the vector dataset. To accomplish this, we propose selecting which samples will be included in the vector dataset using standard methods. These selection methods will choose a subset of elements from the training set to populate the database based on their specific internal procedures. Techniques such as clustering and coreset methods can be applied or adapted to select the most representative samples for each garment.

\textbf{Clustering}: Clustering methods aim to identify subgroups or distributions within a dataset. In this context, we will apply a clustering technique to the representations $v_i$ of all the images of the same category. For each subgroup $C_j$ found, we then obtain the centroid $c_j$:

\begin{equation}
    c_j = \frac{1}{N} \sum v_i \in C_j
\end{equation}

Here, the centres of the clusters represent the different views we want to store in our database. By selecting the centroids, we drastically reduce the images we add to the vector database. It is important to note that most clustering methods group samples based on similarity in a latent space created by a pre-trained model, which makes these methods prone to error due to the feature vector noise. 

\textbf{Coreset}: Coreset selection methods focus on identifying the most informative subset from the full training set that contributes most to the performance of a trained model. Formally, we consider the training set $\mathcal{T}=\{v_i,g_i\}_{i=1}^{N}$, where $v_i\in \mathcal{V}$ denotes a feature vector in the representation space and $g_i\in\mathcal{G}$ represents the corresponding garment categorization as the ground-truth label of $v_i$. The goal of coreset selection is to find a subset $\mathcal{S}$ with $\mathcal{S}\subset\mathcal{T}$ such that a model $F^{\mathcal{S}}_{\theta}(v): \mathcal{V}\rightarrow\mathcal{G}$ trained only on $\mathcal{S}$ will have close generalization performance to the same model $F^{\mathcal{T}}_{\theta}(v)$ that are trained on the whole training set $\mathcal{T}$.

Our approach does not directly train a model for the fashion image retrieval task, making direct model comparison infeasible for some coreset selection methods. To test some Coreset methods, we train a simple auxiliary classifier for garment categorisation. This auxiliary model serves as a proxy for evaluating the effectiveness of different coreset selection strategies.

\subsection{Removing uncharacteristic samples}
\label{sec:rem_out}

\begin{figure}[t]
  \centering
   \includegraphics[width=0.9\linewidth]{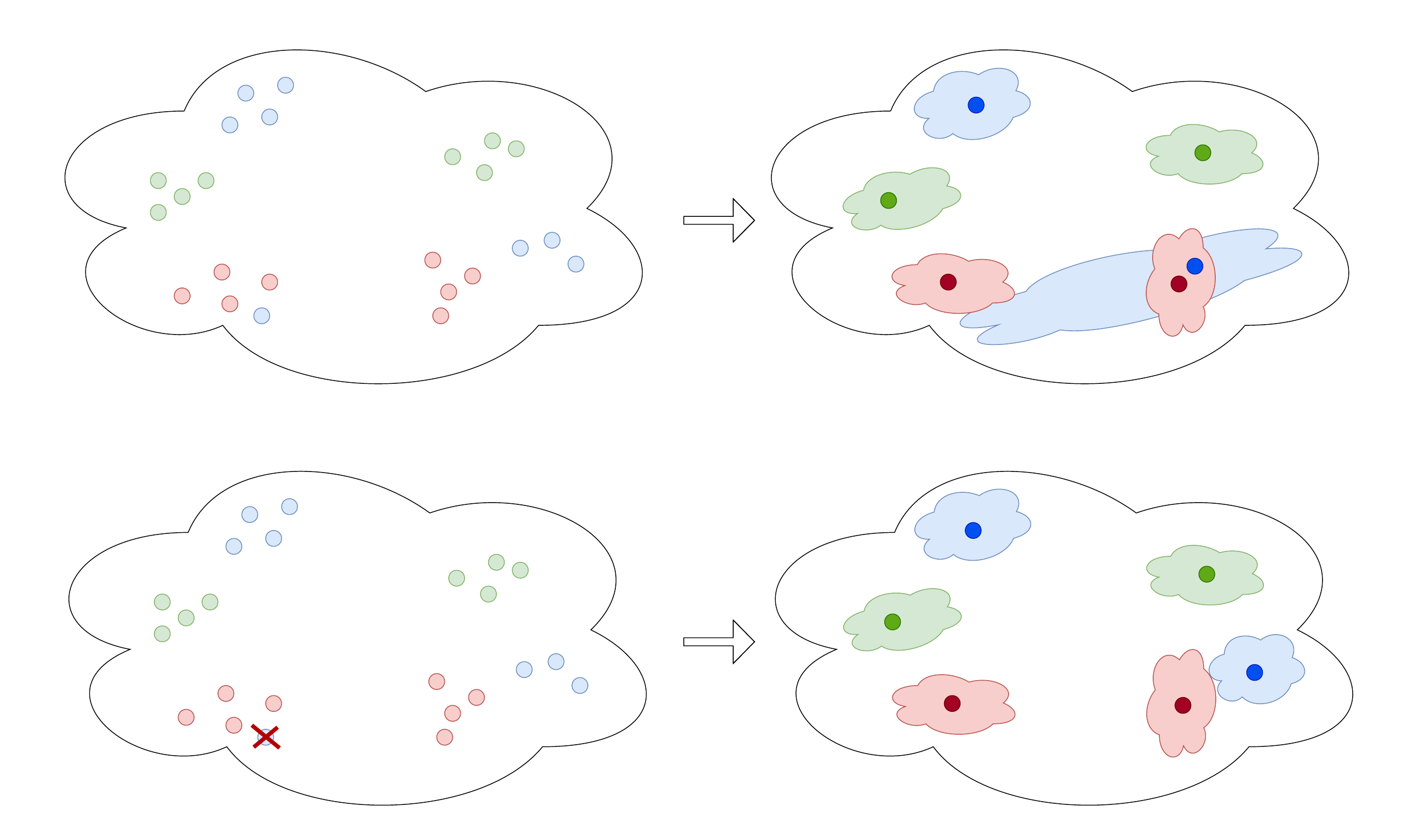}
   \hspace{5mm}
   \caption{Outliers can affect the concept of the sub-distribution, which affects the position of the centroid in the space. As we apply clustering only over the class elements, these affected centroids can easily be close to concepts representing other categories. By removing the non-discriminative, centroids are not distracted by spurious elements.}
   \hspace{150mm}
   \label{fig:outliers_removal}
\end{figure}

Applying the techniques described above to each category allows for the precise identification of the various internal groups within each garment. However, as we will see in the results, the centroids do not consider the similarity between possible subgroups, making it feasible that a centroid represents a category but not discriminative between classes due to their similarity. This may be due to the internal variability and possible noise in the examples. Therefore, we propose removing noisy samples first and then applying a clustering method, as illustrated in Figure \ref{fig:outliers_removal}.

Inspired by the Consistency Score (C-Score) \cite{jiang2021characterizing}, we aim to identify elements highly correlated with samples from different classes. In other words, we want examples where the model may have low confidence in classifying them. However, a major challenge of the C-Score is its high computational cost, which can make it nearly unfeasible in environments characterised by significant data distribution changes. Given these limitations, exploring approximations of the C-Score \cite{jiang2021characterizing, hurtado2023memory} is crucial.

We propose analysing the labels of a given sample's neighbours in the embedding space to obtain the C-Score approximation. Specifically, we calculate the ratio of neighbours that share the same label as the sample. For a given sample $\{x, y\}$, we determine its label homogeneity $H$ in relation to its nearest $N$ neighbors $\{x_i, y_i\}, i=1..N$, according to Equation \ref{eq:score}.

\begin{equation}
    \label{eq:score}
   H(x,y) = \frac{1}{N}\sum \limits_{i=1}^N 1[y = y_i].
 \end{equation}

After obtaining a score for each sample, we can set a threshold $h$ to eliminate samples where $H(x_i) < h$. The intuition behind this method is that samples with low scores are likely surrounded by examples from other classes, thus, cannot be considered as representatives, so if we include these low-scoring samples in the dataset, they might be confused with examples of different garments. Removing these samples enhances the selection process, allowing to focus on more \textit{representative} and \textit{distinctive} samples of the garments.

%%%%%%%%%%%%%%%%%%%%%%%%%%%%%%%%%%%%%%%%%%%%%%%%%%%%%%%%%%%%%%%%%%%%%%%%

\section{Results}

In this section, we present the results we obtained when we used different methods to select examples (Sect. \ref{sec:rep_garment}), then we show how the results change when using the C-Score (Sect. \ref{sec:rem_samples}), and finally, we conclude with ablation to explain some of the decisions made.

\subsection{Experimental Setup}

To evaluate the robustness of our hypotheses, we conducted experiments on three publicly available benchmarks. The first benchmark is Deep Fashion \cite{liuLQWTcvpr16DeepFashion}, using images of the attribute detection task. Using the information of the garment category, we split the dataset into three random train/test sets while preserving the structure and balance of the dataset. For each garment class, we selected a percentage of the images; if a category had fewer than two images, we did not include any from that category in the test set.

For the second benchmark, we use the images from Deep Fashion but now with images of the Consumer-to-Shop (Con2Shop). This benchmark is much more challenging than the previous one, as the images from the same category are usually very different. To represent a real-life environment, we use all images assigned to "shop" as a training set and those assigned to "consumers" as a test.

The final benchmark we consider is Deep Fashion 2 \cite{DeepFashion2}, which provides information for various tasks related to each image, including segmentation and classification. To ensure the benchmarks are comparable, we focus on categorising garments. We use the predefined train/test split.

These benchmarks were chosen over others, such as Fashion 200k \cite{han2017automatic} or MVC \cite{liu2016mvc}, because their images closely resemble real-life situations. The selected benchmarks present challenges such as occlusion, variations in lighting and high variety in the views of a garment. Table \ref{tab:summ_benchmark} provides a summary of the benchmarks.

\begin{table}[]
\caption{Summary of the Benchmarks}
\centering
\begin{tabular}{l|lllll}
\toprule
                        & \# Train  & \# Test   & \# Garm. & |Min| & |Max| \\
\midrule
DF - Attr.    & 16147     & 3853      & 45 & 1 & 3447 \\
DF - Con2Shop & 45392     & 194165    & 16 & 21 & 18265 \\
DF 2          & 76107     & 12684     & 13 & 153 & 15588 \\
\bottomrule
\end{tabular}
\label{tab:summ_benchmark}
\end{table}

\subsection{Baselines}
We primarily focus on comparing clustering and coreset techniques to select the most relevant samples from the dataset. For the clustering methods, we employ kMeans \cite{macqueen1967}, HDBSCAN \cite{campello2015hierarchical}, and CLASSIX \cite{CG24}, implemented in scikit-learn \cite{scikit-learn} and CLASSIX \cite{CG24} Python packages. Regarding coreset techniques, we choose them based on their internal methodologies: Craig, Forgetting, GraNd, Herding, k-Center Greedy, and Uncertainty-Entropy (UE), all of which are implemented in DeepCore \cite{guo2022deepcore}. 

To strengthen our experiments, we test multiple hyperparameters for each method. We mainly focus on the number of clusters generated for the clustering methods. In the case of kMeans, the hyperparameter is straightforward, and we vary the number of clusters between 20 and 100 per garment. In the cases of CLASSIX and HDBSCAN, we explore the cluster sizes that reflect the range of created clusters. We set this range between 1 and 7 based on our empirical results. In the results and figures, we present the best performance observed.

For the Coreset methods, we select a percentage of the full dataset ranging from 10\% to 90\%, increasing with increments of 10\%. For methods that require training, we test a range of learning rates between 0.1 and 0.5. We also utilised category information by allowing or disallowing balanced selection for each garment. All experiments were compared against uniform sampling.

For the experiments that use the C-Score, we range the number of neighbours (N) used between 3 and 50. In the ablation section, we show how different threshold values affect the retrieval performance. 

\subsection{Experiments}

Similar to Section \ref{sec:metho}, we first show the advantages of correctly selecting groups of examples, maintaining performance while decreasing computational cost. We then show the advantages of removing samples with our proposed C-Score approach.

\subsubsection{Garment Representatives}
\label{sec:rep_garment}

\begin{figure*}
  \centering
  \begin{subfigure}{0.33\linewidth}
     \includegraphics[width=0.95\linewidth]{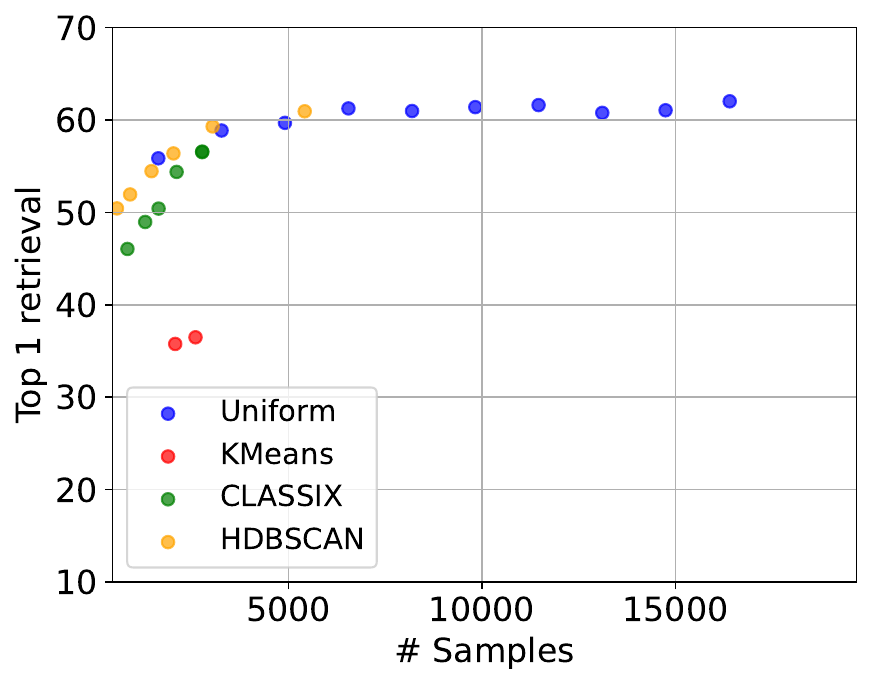}
    \caption{DF - Attr. - Clusters}
    %\label{fig:short-a}
    \hspace{15mm}
  \end{subfigure}
  \hfill
  \begin{subfigure}{0.33\linewidth}
    \includegraphics[width=0.95\linewidth]{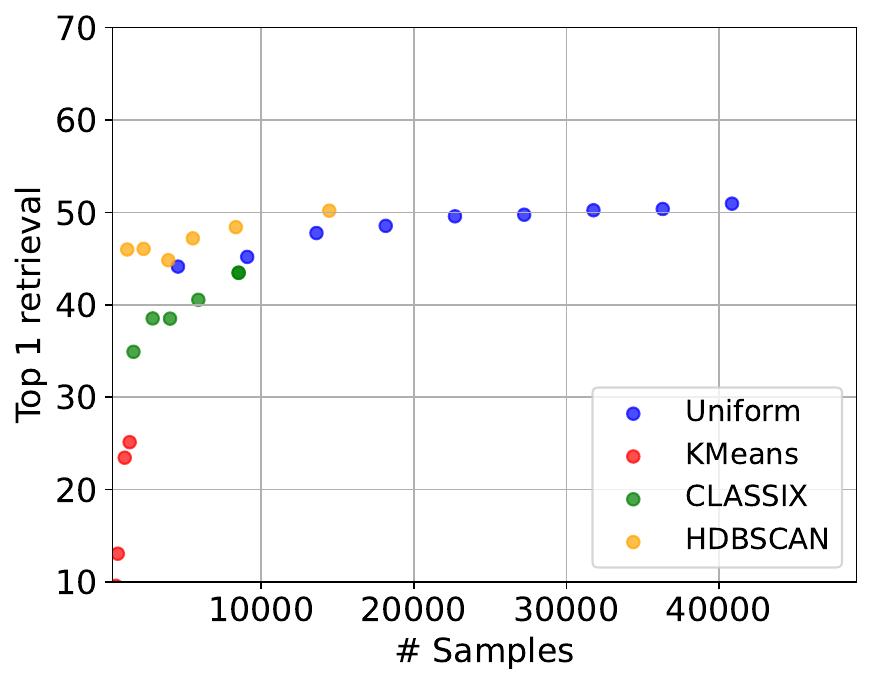}
    \caption{DF Con2Shop - Clusters}
    %\label{fig:short-b}
    \hspace{15mm}
  \end{subfigure}
  \hfill
  \begin{subfigure}{0.33\linewidth}
    \includegraphics[width=0.95\linewidth]{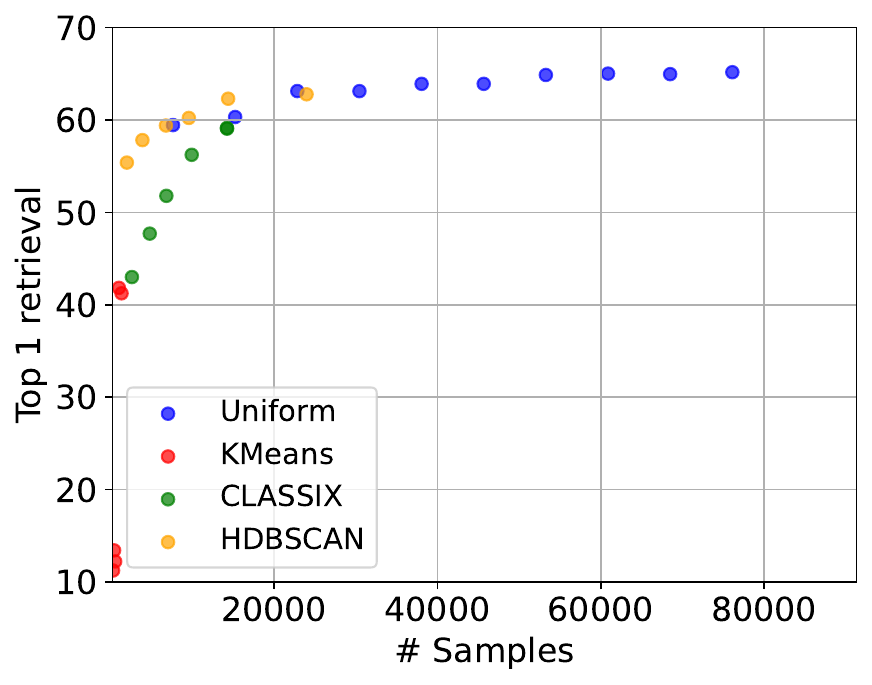}
    \caption{DF 2 - Clusters}
    %\label{fig:short-b}
    \hspace{15mm}
  \end{subfigure}
  
  \begin{subfigure}{0.33\linewidth}
     \includegraphics[width=0.95\linewidth]{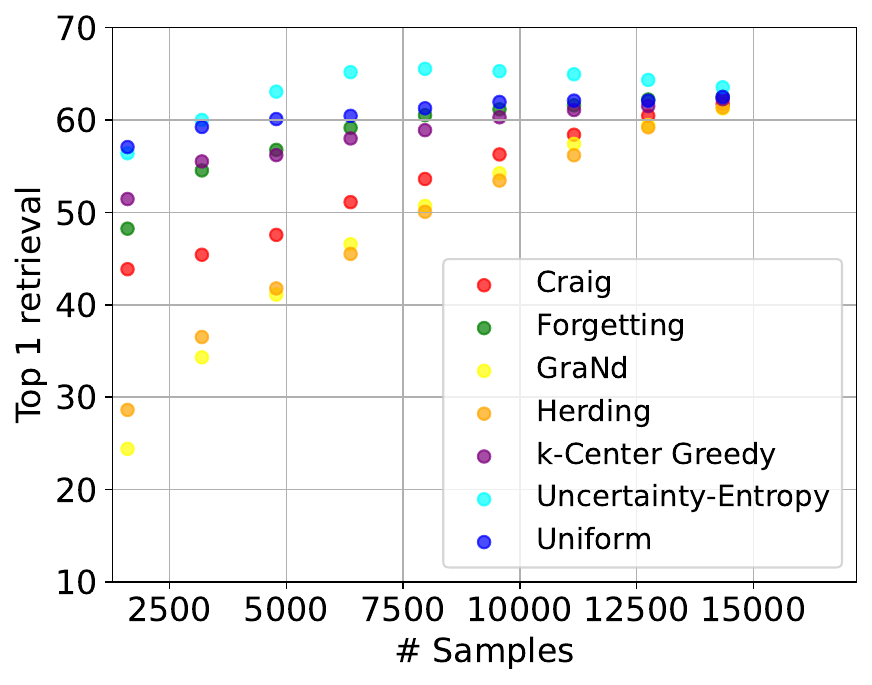}
    \caption{DF - Attr. - Coreset}
    %\label{fig:short-a}
  \end{subfigure}
  \hfill
  \begin{subfigure}{0.33\linewidth}
    \includegraphics[width=0.95\linewidth]{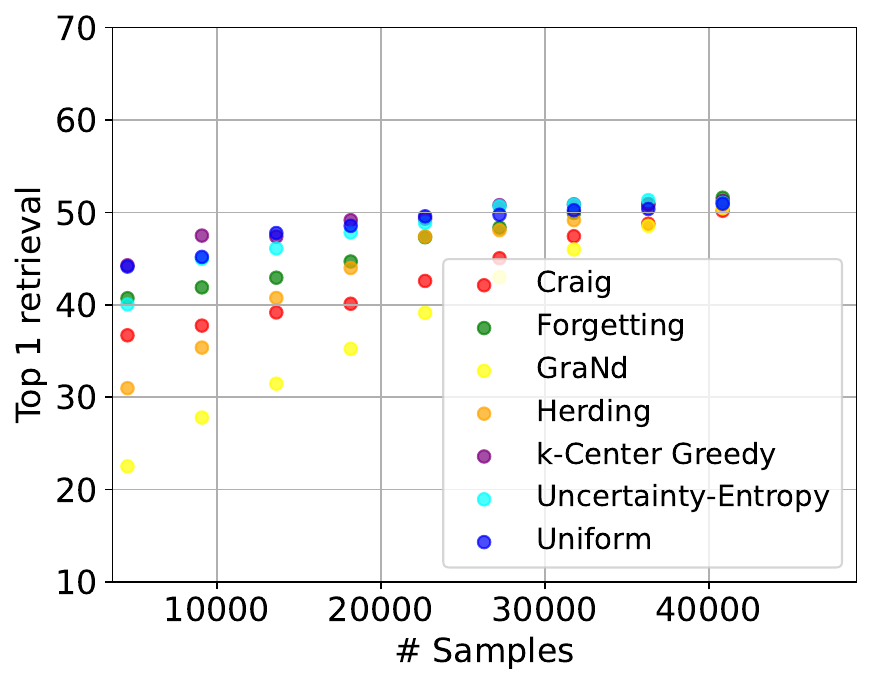}
    \caption{DF Con2Shop - Coreset}
    %\label{fig:short-b}
  \end{subfigure}
  \hfill
  \begin{subfigure}{0.33\linewidth}
    \includegraphics[width=0.95\linewidth]{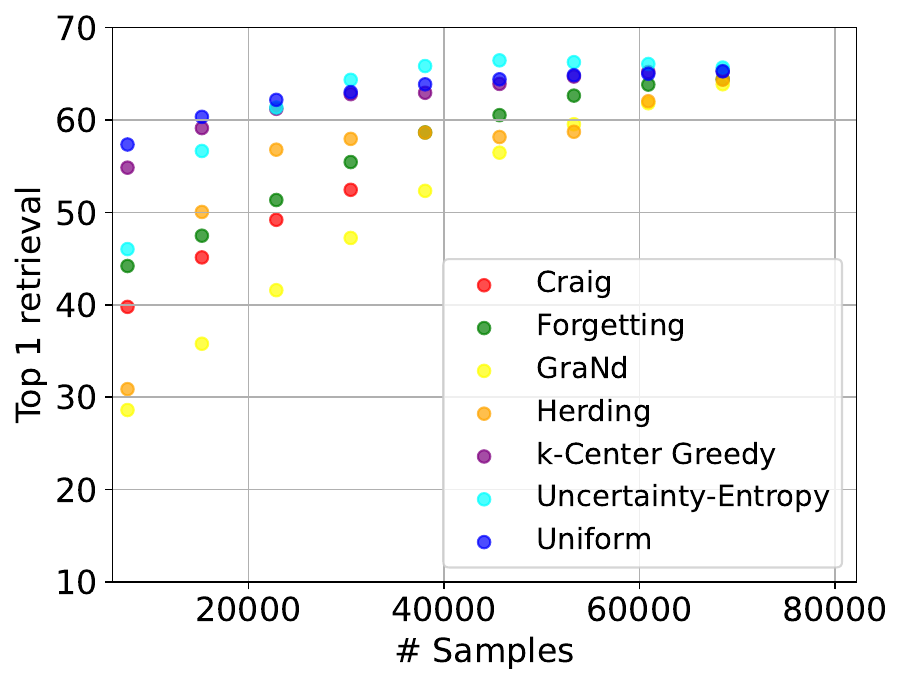}
    \caption{DF 2 - Coreset}
    %\label{fig:short-b}
  \end{subfigure}
  \hspace{5mm}
  \caption{Performance obtained by different methods compared to a random selection. The first row shows the accuracy with clustering-based methods, and the second row shows results with Coreset-based methods. Each column is a different benchmark.}
  \label{fig:representation_space}
  \hspace{5mm}
\end{figure*}

The second-hand fashion market exhibits diversity in terms of data availability and image distribution. This can be observed in the selected benchmarks as shown in Table \ref{tab:summ_benchmark} and Figure \ref{fig:diversity_garms}. The table highlights a considerable discrepancy between the minimum and maximum data available for each garment. Meanwhile, the figure illustrates how visually different images of the same garment can be, making it challenging for even humans to identify matching pairs accurately. This variability can negatively impact performance, as the selection of images may not be optimal or may lack sufficient representation of the garments. 

Using clustering techniques to select images reveals an interesting pattern, as shown in the first row of Figure \ref{fig:representation_space}. One notable observation is the poor performance of kMeans and CLASSIX compared to uniform selection. Among the methods tested, HDBSCAN performed the best, with similar performance with uniform selection. The exception was DF-Con2Shop, in which HDBSCAN outperformed random selection by 2\% while using fewer examples. The poor performance of kMeans highlights the necessity of filtering out elements not represented by any cluster. On the other hand, both CLASSIX and HDBSCAN carry out this filtering process internally, which likely contributes to their stronger performance.
It is also important to note that we only select the cluster centres as representative elements, which limits the number of examples we can sample.

For the case of Coreset-based methods, we see a different trend than Clustering-based approaches, as seen in the second row in Figure \ref{fig:representation_space}. A random selection tends to perform better when a few examples are selected. However, as we add items to the database, UE improves its performance, even outperforming the random selection with 30-70\% of the data. As expected, all methods converge to the same point as we increase the percentage. 

HDBSCAN and UE improve search efficiency by reducing the number of items stored in the database without compromising retrieval accuracy. The intuition behind these improvements is that both methods effectively disregard low-consistency samples. HDBSCAN creates clusters only from highly dense sub-distributions that better represent the garments. On the other hand, UE selects elements with high entropy (low uncertainty) that are primarily located in areas of the distribution where samples of the same garment are concentrated. 

Both methods successfully address different levels of data availability. HDBSCAN excels at choosing a minimal set of samples (10-20\%), while UE surpasses random selection when selecting 30-70\% of the data. This demonstrates that excluding certain samples can enhance the retrieval of similar samples and improve Top 1 retrieval times.

Building on this intuition, we integrate the core concepts of both methods by first eliminating noisy samples and then obtaining centres. The balance between optimal performance and minimal search time, as illustrated in Figure \ref{fig:elapsed_time}, is heavily influenced by the specific scenario at hand. Ideally, we aim to achieve good performance while minimising the number of examples added to the database.

\subsubsection{Removing uncharacteristic samples}
\label{sec:rem_samples}

\begin{figure*}
  \centering
  \begin{subfigure}{0.33\linewidth}
     \includegraphics[width=0.85\linewidth]{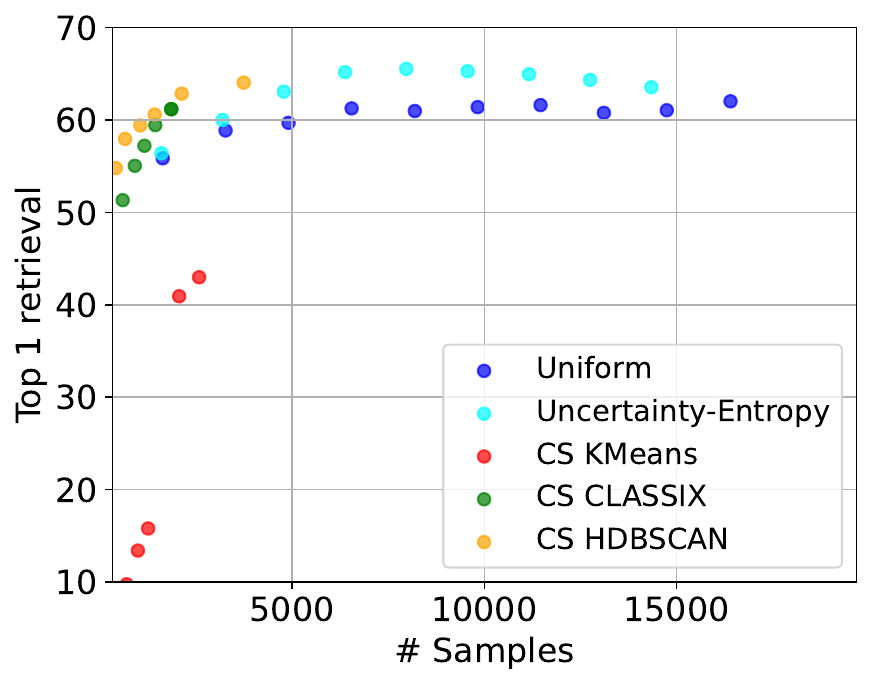}
    \caption{DF - Attr. - C-Score}
    %\label{fig:short-a}
  \end{subfigure}
  \hfill
  \begin{subfigure}{0.33\linewidth}
    \includegraphics[width=0.85\linewidth]{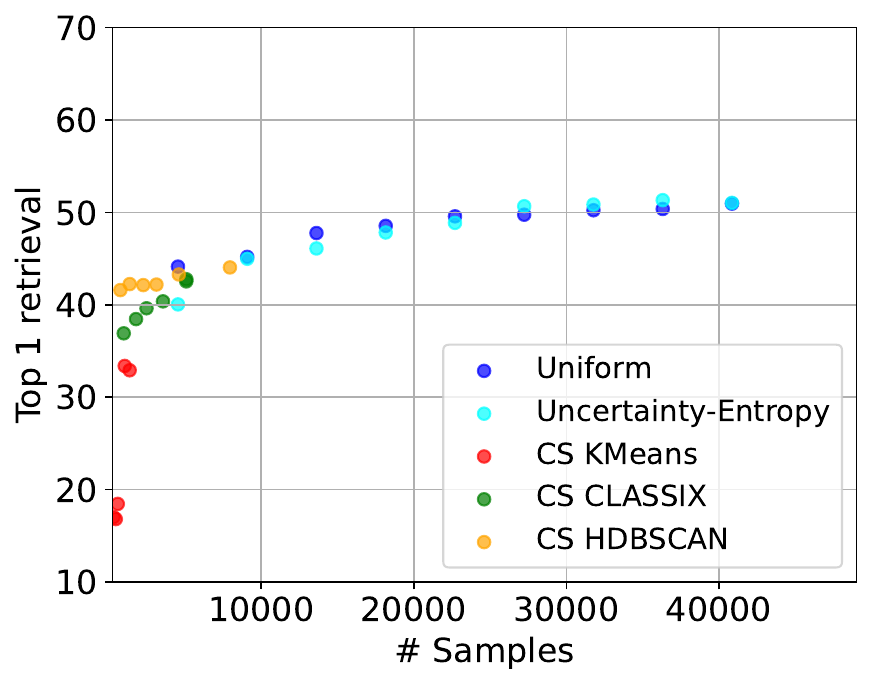}
    \caption{DF Con2Shop - C-Score}
    %\label{fig:short-b}
  \end{subfigure}
  \hfill
  \begin{subfigure}{0.33\linewidth}
    \includegraphics[width=0.85\linewidth]{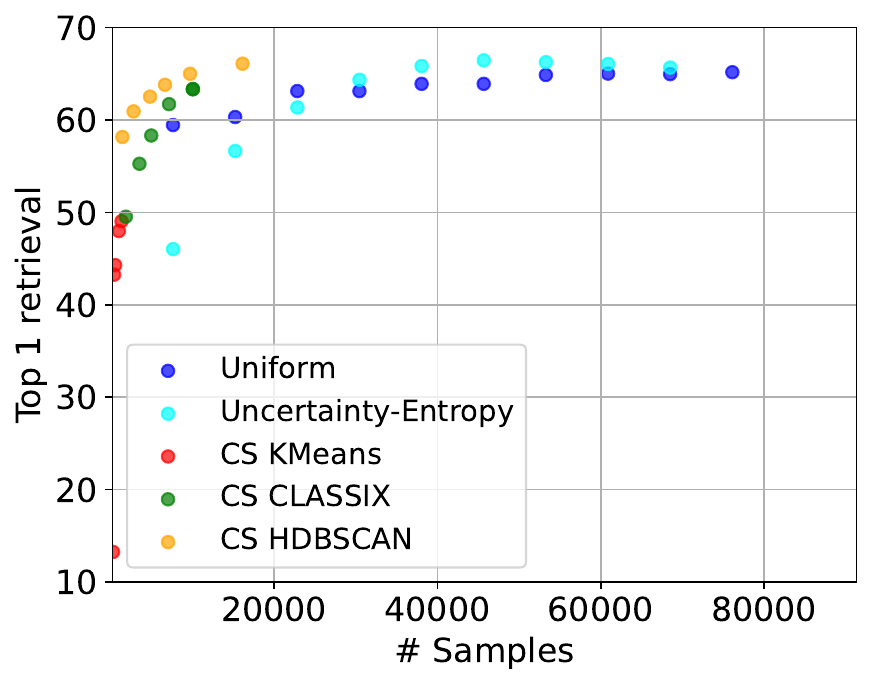}
    \caption{DF 2 - C-Score}
    %\label{fig:short-b}
  \end{subfigure}
  \hspace{5mm}
  \caption{On DP - Att. and DF 2, the performance of all methods increases as we apply the removal using the C-Score, with HDBSCAN and CLASSIX even outperforming uniform sampling. However, on DP - Con2Shop, the performance of all methods decrease.}
  \label{fig:representation_space_c_score}
\end{figure*}

Based on the findings from the previous section, the current hypothesis proposes that removing low-confidence samples may enhance performance for two reasons: (1) it helps to identify clusters unaffected by noise, and (2) these noisy samples are typically uncharacteristic and may not effectively distinguish between classes. However, identifying these samples in highly noisy distribution, like the second-hand market, can be challenging. The uniqueness of each image makes it difficult to determine which samples are truly noisy and which are valid representatives of their classes, albeit less common.

The intuition is that the most harmful samples create sub-groups similar to those of other categories. In other words, samples that are generic or resemble other garments—such as an image of blue jeans lacking any distinctive features. This issue would have a negative influence on the creation of centroids in clustering-based methods, as it would lead to centroids that do not accurately represent the actual distribution of the test data. As a result, this may decrease generalizability and ultimately deteriorate the performance.

Our method removes outliers by filtering samples surrounded by instances of the other classes and far from samples of their category. This denoising step reduces the overlap of representatives of different garments, allowing the clustering method to focus on discriminative samples.
The results of applying the C-Score removal are presented in Figure \ref{fig:representation_space_c_score}.
 
In most cases, removing examples before applying clustering helps increase the performance and validate our hypothesis. This improvement can be seen especially in DF Attr. and DF 2, where CS-HDBSCAN (HDBSCAN with C-Score) outperforms random selection and performs similarly to UE but using far fewer examples.

\subsection{Ablations}

\textbf{Model Selection}
As mentioned in the Related Work section, employing pre-trained models to create visual representations has become a standard practice in various applications that involve searching for similar images, especially when domain-specific large pre-trained models exist. There are several models specifically fine-tuned for the fashion domain. However, the performance of models trained on millions of generic data points can be just as relevant as those specialized on domain-specific data.

In this comparison, we examine two general visual models, CLIP and ViT, alongside three specialised fashion models. The main distinction between CLIP and ViT is that CLIP has been trained using a multimodal approach, while ViT focuses exclusively on visual data. Table \ref{tab:models_select} presents the performance metrics of each model, displaying the top 1, top 5, and top 20 retrieval accuracy results in the DF - Attri benchmark. This metric indicates the probability that at least one item from the same class is included in the retrieved elements. As anticipated, the domain-specific models outperform the general ones, with Marqo-FashionCLIP being the top performer, which we utilize in all experiments.

\begin{table}[]
\caption{Performance obtained when generating the feature vector with different models in DF - Attr.}
\centering
\begin{tabular}{l|lll}
\toprule
                                                & Top1  & Top 5 & Top 20 \\
\midrule
ViT \cite{dosovitskiy2020image}                 & 52.48 & 77.81 & 91.00  \\
CLIP \cite{radford2021learning}                 & 58.32 & 83.49 & 93.83  \\
Fashion-CLIP2.0 \cite{Chia2022}                 & 59.39 & 83.68 & 93.91  \\
Open Fashion-CLIP \cite{cartella2023open}       & 60.18 & 83.91 & 93.98  \\
Marqo-FashionCLIP \cite{Marqo_fashionclip}      & \bf{62.66} & \bf{85.24} & \bf{94.40}  \\
\bottomrule
\end{tabular}
\label{tab:models_select}
\end{table}

\noindent\textbf{C-Score hyper-parameters}
Our approximation of the C-Score includes two hyperparameters. The first hyperparameter is the number of neighbours we consider when calculating the consistency value for each sample. A higher value enables us to account for a larger number of neighbours, which can be beneficial when we have many samples from different classes in close proximity. Conversely, a smaller value helps mitigate issues that arise when numerous distinct categories are present.

The second hyperparameter value represents the threshold to identify noisy samples for removal from the database. There is a trade-off involved: on one hand, we aim to eliminate as many noisy samples as possible; however, if we remove too many, we risk discarding potentially relevant but uncommon samples.

\begin{figure*}
  \centering
  \begin{subfigure}{0.33\linewidth}
     \includegraphics[width=0.95\linewidth]{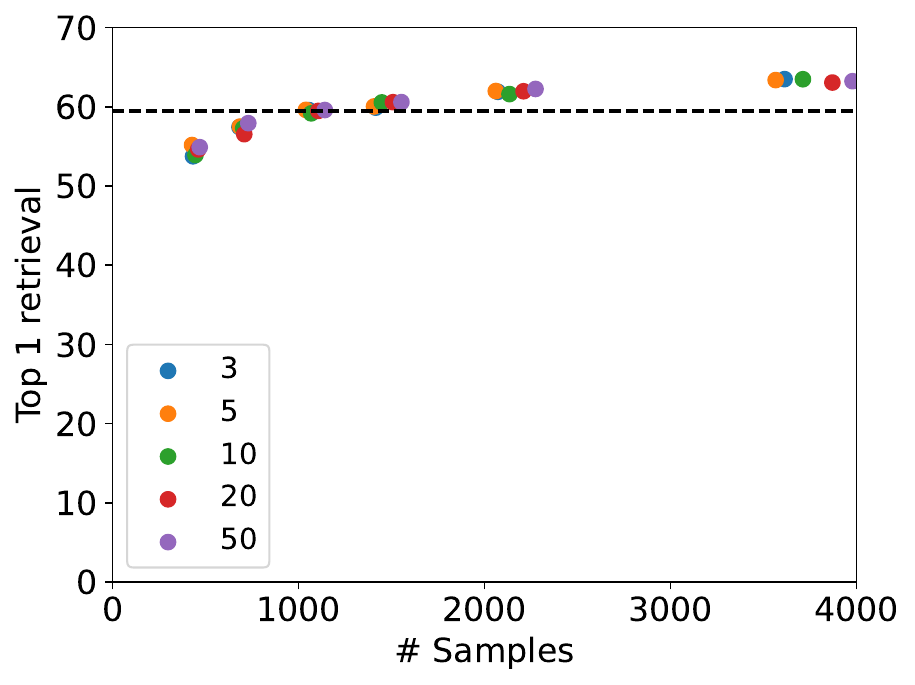}
    \caption{\# Neighbours - Threshold: 0.2}
    %\label{fig:short-a}
    \hspace{15mm}
  \end{subfigure}
  \hfill
  \begin{subfigure}{0.33\linewidth}
    \includegraphics[width=0.95\linewidth]{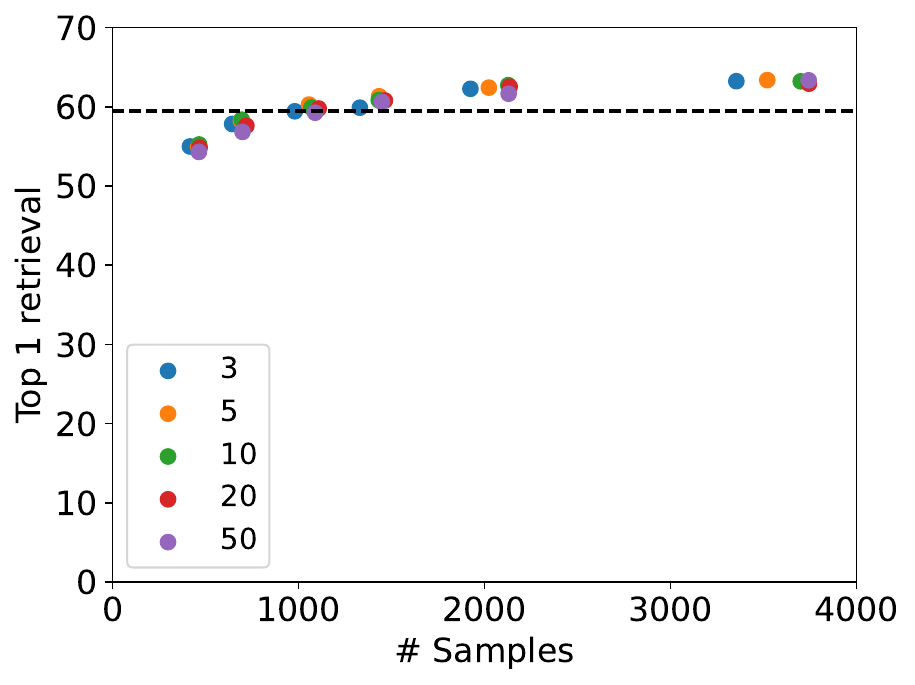}
    \caption{\# Neighbours - Threshold: 0.4}
    %\label{fig:short-b}
    \hspace{15mm}
  \end{subfigure}
  \hfill
  \begin{subfigure}{0.33\linewidth}
    \includegraphics[width=0.95\linewidth]{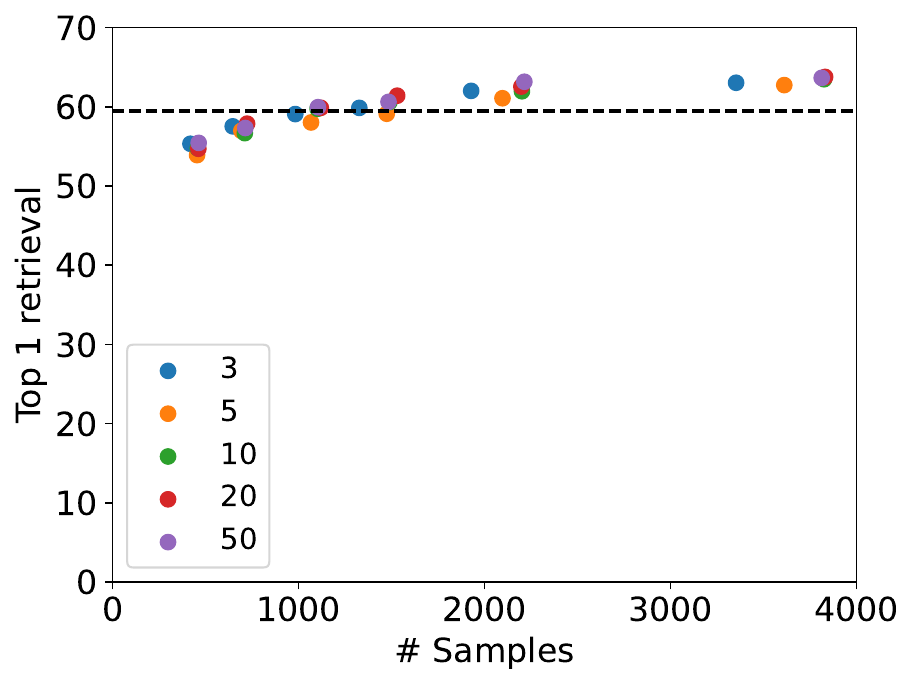}
    \caption{\# Neighbours - Threshold: 0.6}
    %\label{fig:short-b}
    \hspace{15mm}
  \end{subfigure}
  
  \begin{subfigure}{0.33\linewidth}
     \includegraphics[width=0.9\linewidth]{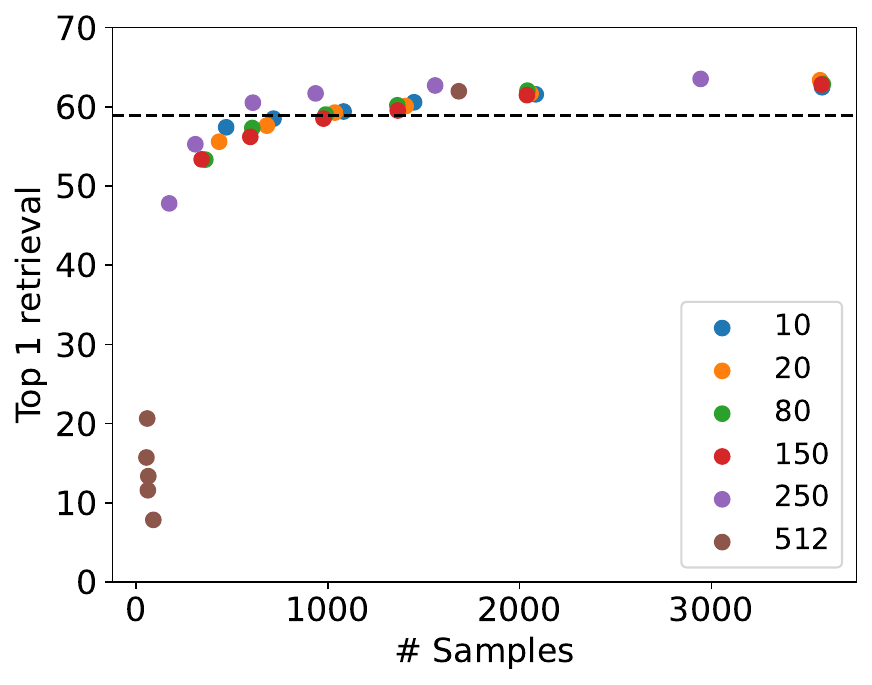}
    \caption{\# Components - Threshold: 0.2}
    %\label{fig:short-a}
  \end{subfigure}
  \hfill
  \begin{subfigure}{0.33\linewidth}
    \includegraphics[width=0.9\linewidth]{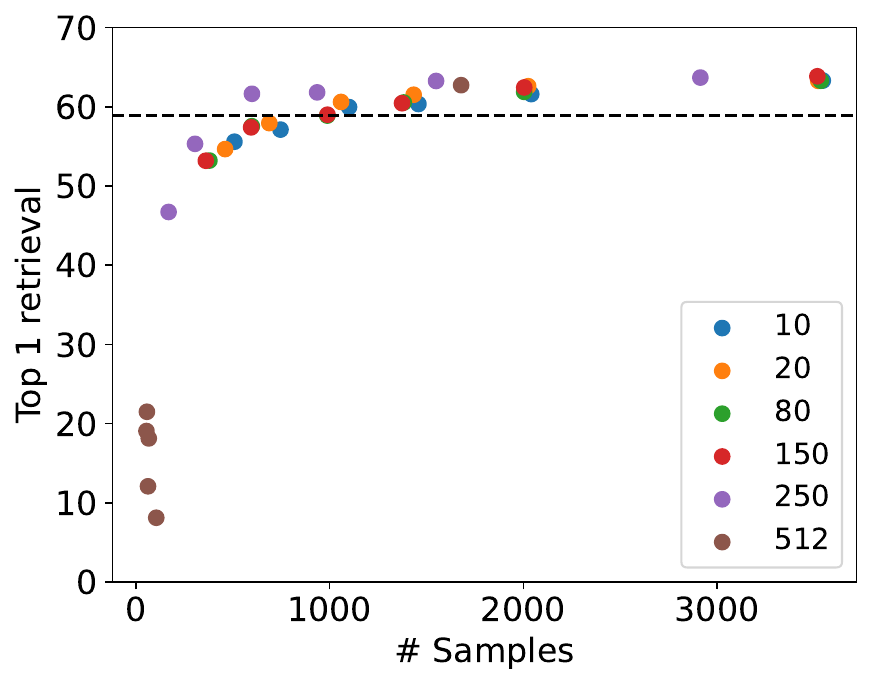}
    \caption{\# Components - Threshold: 0.4}
    %\label{fig:short-b}
  \end{subfigure}
  \hfill
  \begin{subfigure}{0.33\linewidth}
    \includegraphics[width=0.9\linewidth]{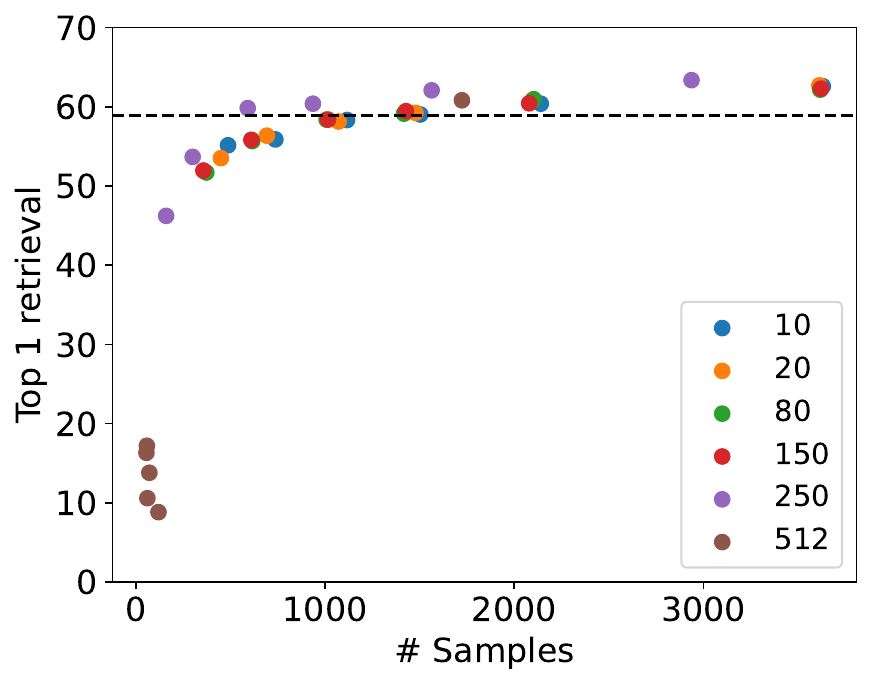}
    \caption{\# Components - Threshold: 0.6}
    %\label{fig:short-b}
  \end{subfigure}
  \hspace{5mm}
  \caption{Ablation of the hyperparameters of the proposed method. The first row presents results with varying values of the number of neighbours, while the second row displays results from different numbers of components used in the UMAP application.}
  \label{fig:abla_hp}
\end{figure*}

The first row in Figure \ref{fig:abla_hp} shows the performance of HDBSCAN using various values for the number of neighbours and different thresholds in the DF - Attr benchmark. The results indicate that the method is robust when both the number of neighbours and the thresholds are altered. The horizontal line in the figure represents the best value achieved when applying HDBSCAN without the C-Score.

We also explore the performance of the proposal when UMAP is used to select various components before removing samples. The second row of Figure \ref{fig:abla_hp} shows how performance varies with different thresholds. The methods tend to be robust regarding the number of components used in the feature vector; however, when UMAP is not applied (and set to 512 components), the method performs poorly. These results underscore the importance of having effective representations to achieve successful retrieval.

\begin{figure}[]
  \centering
  \begin{subfigure}{0.95\linewidth}
     \includegraphics[width=0.95\linewidth]{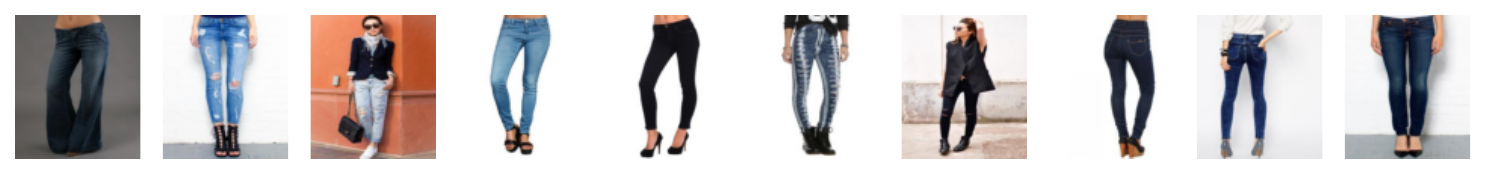}
     \vspace{-2mm}
     \caption{Uniform Sampling}
     \vspace{4mm}
  \end{subfigure}
  \begin{subfigure}{0.95\linewidth}
     \includegraphics[width=0.95\linewidth]{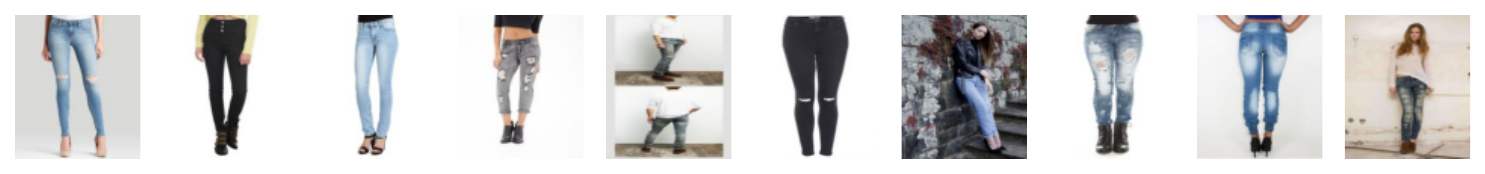}
     \vspace{-2mm}
     \caption{HDBSCAN Sampling}
     \vspace{4mm}
  \end{subfigure}
  \begin{subfigure}{0.95\linewidth}
     \includegraphics[width=0.95\linewidth]{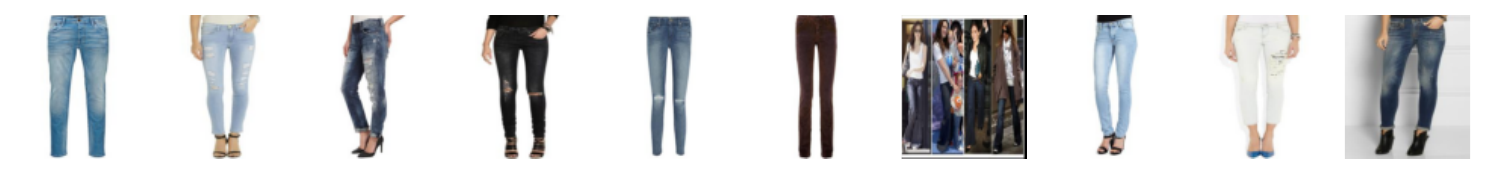}
     \vspace{-2mm}
     \caption{Uncertainty - Entropy Sampling}
     \vspace{4mm}
  \end{subfigure}
  \begin{subfigure}{0.95\linewidth}
     \includegraphics[width=0.95\linewidth]{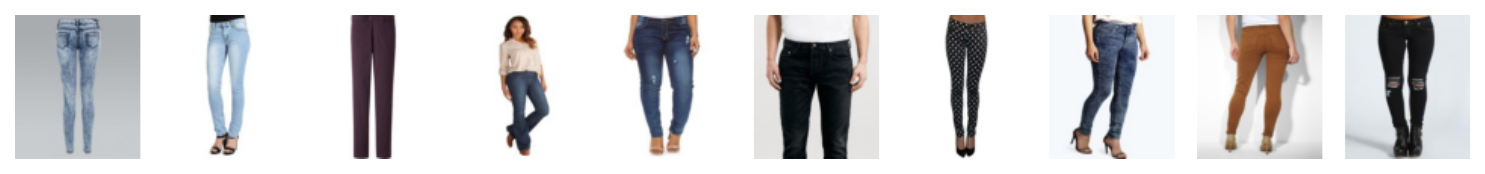}
     \vspace{-2mm}
     \caption{HDBSCAN with C-Score Sampling}
     \vspace{4mm}
  \end{subfigure}
   \caption{Centres of a category using different methods.}
   \vspace{5mm}
   \label{fig:sample_center}
\end{figure}

\noindent\textbf{Centres selected}
One way to understand the results obtained in the paper is by visually analysing the centroids produced by different methods. Figure \ref{fig:sample_center} illustrates the centres selected using various techniques. Uniform sampling from the dataset yields a good selection of centroids with diverse colours and backgrounds. However, it also includes clear outliers, such as an image of a person standing against an orange background. Similar patterns can be observed when using HDBSCAN and Uncertainty methods, which also select images featuring unique backgrounds. While these images may effectively represent garments with distinctive characteristics, they likely represent elements that do not capture a complete category distribution.

By removing noisy elements from the dataset, we obtained the centres shown in the last row of Figure \ref{fig:sample_center}. These images showcase a diverse range of colours, lighting conditions, and angles. The advantage of this selection is that it excludes samples with unique backgrounds or features unique to one image. 

This process involves a trade-off between how representative an example can be of a category and how well it reflects the actual distribution. While individual examples may be very representative, they can be less helpful in building the database if they do not closely resemble other examples from the same category, creating more harm than benefit. Conversely, sub-distributions within a category can capture a larger number of examples, highlighting specific characteristics of particular garments, which is valuable for enriching the vector database.

\section{Impact}
As part of the proposal evaluation, we conducted evaluations using images from a single brand on an internal dataset provided by TRUSS\footnote{\url{https://www.truss.fashion/}}. One of the company's goals is to retrieve the most similar listing created by sellers in response to an image uploaded by a user. In this context, we define a listing as a collection of one or more images. With this information, they provide additional details like textual descriptions and historical statistics, like selling prices.

The main challenge they faced was the exponential growth of their database. To address this challenge, their initial solution only considered one image per category to populating the database. With this approach, they achieved a performance rate of only 21\% for the top retrieval result. While the accuracy improved as the number of retrieved elements increased, this led to complications in visualising the results. If they increased the number of samples in the dataset, the performance could rise to 57\% by utilising all available images. However, this would also lead to an increase in computational costs, both for storage and retrieval, as the number of images would increase from 567 to 38707. Using the method described in the paper, we reduced the number of images in the database to 8,754, achieving a performance of 68\%.

Future directions of this collaboration will focus on testing retrieval performance by creating feature vectors through Parameter-Efficient Fine-Tuning (PEFT) models, such as LoRA. Additionally, we will explore methods to combine knowledge from different LoRA weights based on their characteristics.

\section{Conclusion}
There is a natural trade-off between the number of samples added to a vector database and the performance of retrieving the most relevant samples. More samples mean better performance but with a higher computational cost. In this paper, we tackled the challenge of decreasing the computational cost of retrieving relevant images by reducing the samples used to create the vector database without decreasing the retrieval performance. We employ clustering and coreset methods to select the representative samples to populate the vector database. However, despite improvements in efficiency (same performance but with less data), they fail to surpass the performance of a random selection in part of the benchmarks. For this reason, we proposed a method that removes outliers and then applies clustering methods. We demonstrated that this significantly improves retrieval performance using only 10\% of the dataset. Beyond empirical gains, our pipeline aligns with sustainability goals by curbing unnecessary comparisons and model retraining as databases grow. Future work will explore dynamic selection strategies under concept drift, adaptive thresholds for C-Score filtering, and extensions to multi-modal embeddings, such as integrating textual metadata and PEFT. We also plan to investigate online selection algorithms that update the representative vectors in real-time as new garments enter the market to enhance efficiency and retrieval performance further.

%%%%%%%%%%%%%%%%%%%%%%%%%%%%%%%%%%%%%%%%%%%%%%%%%%%%%%%%%%%%%%%%%%%%%%%%

%%% Use this environment to include acknowledgements (optional).
%%% This will be omitted in doubleblind mode.

\begin{ack}
This work is funded by Innovate UK as part of the project ``Multimodal Catalogue Search For Second Hand Apparel Valuations'' (10101553). We would like to thank Dan Edwards, Jack Cardwell, William Parker and Jack Zuliani for their technical support and discussion.
\end{ack}

%%%%%%%%%%%%%%%%%%%%%%%%%%%%%%%%%%%%%%%%%%%%%%%%%%%%%%%%%%%%%%%%%%%%%%%%

%%% Use this command to include your bibliography file.
%\newpage
\bibliography{m1859}

\end{document}